# Adaptive Gated Deepfake Detection for Low-Resolution and Resource-Constrained Environments

Vaishnavi Sen
*Department of Computer Science*
*California State University, Northridge*
California, USA
0009-0009-2245-9596

Cody Laurie
*Department of Computer Science*
*California State University, Northridge*
California, USA
0009-0003-2740-964X

Rashida Hasan
*Department of Computer Science*
*California State University, Northridge*
California, USA
0000-0002-6231-8116

***Abstract*—Deepfake detection models often rely on high-quality inputs, fixed inference paths, and computationally expensive architectures, limiting their use in low-resolution and resource-constrained settings. This paper proposes AdaGate-DF, an adaptive gated deepfake detection framework that uses image-quality cues to route samples through a dual multi-exit system so high-quality images can exit earlier and save compute. We evaluated AdaGate-DF against MaD-CoRN, DefakeHop++, and ShuffleNetV2 on two benchmark datasets (Celeb-DF and FaceForensics++) under multiple configurations to test image resolution dependence and training and inference efficiency. On Celeb-DF, AdaGate-DF achieves an AUC of 0.9370, outperforming MaD-CoRN and DefakeHop++ while maintaining a low inference latency. Resolution-based testing shows consistent improvement as input resolution increases, reaching an AUC of 0.9708 at $384 \times 384$. The FaceForensics++ results highlight that AdaGate-DF remains effective under class imbalance, following competitive results with evaluated models. Overall, AdaGate-DF demonstrated a practical balance between detection performance, uncertainty-aware prediction, and computational efficiency for variable-quality deepfake detection.**

## I. Introduction

The rapid improvement of generative models has made manipulated facial media, which is increasingly realistic and difficult to distinguish from authentic content. As a result, deepfake detection has become a major problem for digital forensics, media verification, and maintaining online trust. Benchmark datasets such as FaceForensics++ (FF++) [1] and Celeb-DF [2] have supported progress in this area by providing standardized evaluation settings for manipulated facial media. Recent literature also shows that deepfake detection has expanded in multiple research directions, including spatial artifact learning, frequency-domain analysis, lightweight architectures, low-quality image detection, and multimodal feature fusion [3].

Despite this progress, detecting deepfakes under low-resolution and resource-constrained conditions remains challenging. In real-world settings, images and video frames are often compressed, resized, blurred, or captured at inconsistent resolutions. These degradations can reduce or distort fine-grained facial artifacts that are useful for detecting manipulation. Prior work has therefore studied low-quality and low-resolution deepfake detection through optimized preprocessing, machine learning-based low-resolution analysis, and resolution-aware feature extraction [4]–[7]. Other studies have explored super-resolution as a preprocessing strategy for improving low-quality deepfake detection, showing that image enhancement can recover useful visual detail but may also introduce preprocessing-dependent artifacts [8], [9].

Frequency-domain analysis has become an important direction for robust deepfake detection because manipulation and compression artifacts may remain visible outside the RGB domain. The Discrete Cosine Transform (DCT) [10] is often used to provide a compact representation of frequency information and has been widely used in image processing. Deepfake detection methods have used DCT, wavelet, and broader frequency-aware representations to identify high-frequency inconsistencies and manipulation artifacts that may not be easily captured from RGB inputs alone [11]–[14]. Multi-modal and multi-branch methods further support this direction by combining complementary cues from spatial, frequency, and attention-based representations [15]–[17].

Nevertheless, deepfake detection presents continuous challenges in computational efficiency. Many deepfake detectors process every input using the same fixed inference path, regardless of image quality or sample difficulty. This approach is often inefficient for real-world deployment, particularly in settings where large volumes of images and video frames must be processed in real time under limited computational resources, such as social media. As a result, lightweight model design has emerged as a promising direction for real-world deepfake detection, with prior work exploring efficient CNNs, convolutional reservoir networks, multi-feature fusion, and application-specific lightweight detectors [18]–[21]. Low-quality video detection methods have also examined multi-scale and branch-based mechanisms to handle degraded inputs [22], but still rely on the fixed inference behavior irrespective of the image quality. The research usually focuses on one of the challenges at a time, often leading to complex architecture for low resolution images or not considering the degraded resolution of the images when focused on making the architecture lightweight.

To address these challenges, this work proposes **AdaGate-DF**, an adaptive gated deepfake detection framework for low-resolution and resource-constrained environments. AdaGate-DF uses a dual-branch representation that combines RGB spatial features with frequency-domain features obtained through DCT. The framework also incorporates a multi-exit gated inference mechanism that routes samples through fast, medium, or full inference paths based on resolution and image-quality cues. This adaptive design is motivated by early-exit inference, where samples can exit at different network depths to reduce unnecessary computation [23]. In contrast to fixed-depth detection pipelines, AdaGate-DF adjusts the inference path based on input quality, allowing simpler samples to be processed efficiently while reserving deeper analysis for more difficult or degraded inputs.

We evaluated AdaGate-DF on Celeb-DF and FF++ and compared it with three state-of-the-art models, including lightweight and frequency-aware detection approaches. The evaluation considers the overall detection performance, the uncertainty-aware prediction behavior, resolution-based robustness, and computational efficiency in terms of training time and inference latency. The Uncertainty rate along with accuracy, AUC, precision, recall, and F1-score allows for a more reliable evaluation by overcoming the high-confident misclassifications. By examining these factors together, this work focuses on the practical requirements of deploying deepfake detectors for variable-quality and low-resource environments.

The main contributions of this work are summarized as follows:

- We propose **AdaGate-DF**, a dual-branch adaptive gated framework that integrates RGB and frequency-domain representations with a multi-exit inference mechanism for low resource and quality-aware deepfake detection.
- We evaluate the proposed approach on two benchmark datasets, Celeb-DF and FF++, and compared its performance with state-of-the-art methods, which additional evaluation for resolution based performance.
- We incorporate uncertainty-aware evaluation to compare how often each model produces low-confidence predictions, providing additional insight beyond standard accuracy-based metrics.
- We analyze computational efficiency by comparing training time and inference latency across the evaluated models, with an emphasis on practical deployment in low-resource environments.

**Reproducibility:** To support future researchers, all of our code, dataset sources, preprocessing, and evaluation scripts are available at https://github.com/CodeGhost157/AdaGate-DF.

## II. Related Work

Deepfake detection is commonly formulated as a binary classification task that distinguishes authentic facial media from manipulated content. Benchmark datasets such as FF++ and Celeb-DF have supported standardized evaluation in different types of manipulation, compression settings, and visual conditions [1], [2]. Early detectors relied mainly on CNN-based spatial artifact learning, including Xception-style models and compact architectures such as MesoNet [24], [25]. Later, efficient architectures emerged, including EfficientNet, MobileNetV3, and ShuffleNetV2, improving the balance between model size and detection accuracy [18], [26], [27]. However, these methods generally use fixed inference paths and can be sensitive to changes in image quality, compression, and resolution.

Therefore, Low-resolution and low-quality detection has become an important research direction. In real-world settings, deepfake images and video frames may be repeatedly resized, compressed, down-sampled, or blurred before being uploaded to social media. These operations can suppress fine-grained facial artifacts and make spatial-only detection less reliable. Degraded deepfake inputs have been addressed through optimized preprocessing, low-resolution classification pipelines, multi-scale DCT features, and lightweight frequency-domain CNNs [4]–[7]. Super-resolution has also been used to enhance low-quality inputs before detection [8], [9]. These studies show that resolution and preprocessing strongly influence detector performance. However, most focus on improving input quality or feature representation rather than adapting computation to the difficulty of each sample.

Manipulation, compression, and resizing artifacts may remain visible outside the RGB domain, even when they are difficult to detect spatially. As a result, frequency-domain methods have been used to capture frequency level inconsistencies in manipulated faces through DCT, wavelet, and broader spectral representations [10]–[14]. Multi-modal and multi-branch methods further combine complementary RGB, frequency, and attention-based cues to improve robustness under challenging conditions [15]–[17]. While these methods improve representation quality, they often increase model complexity and typically retain a fixed inference pipeline.

Several works have also focused on lightweight or efficient deepfake detection. MaD-CoRN, DefakeHop++, compact CNNs, and multi-feature fusion methods reduce computational cost compared with heavier deep models [19]–[21], [28]. Complementary work on degraded video detection, such as BZNet, uses multi-scale branch processing to improve robustness under low-quality conditions [22]. Although these approaches improve efficiency, they generally use a fixed computational pathway for all inputs. This limits their ability to save computation on easier samples while applying more rigorous analysis to difficult or ambiguous cases. Such adaptive inference is important for low-resource deployment, where large volumes of images or video frames must be processed efficiently without sacrificing reliability.

Adaptive inference addresses this limitation by assigning different network depths to different samples, reducing unnecessary computation for easier inputs while preserving greater capacity for difficult or ambiguous cases. BranchyNet introduced early-exit classifiers that allow confident samples to exit earlier while difficult samples continue through deeper layers [23]. However, most early-exit methods rely primarily on prediction confidence rather than explicit image-quality cues. In low-resolution deepfake detection, quality features such as resolution, blur, sharpness, compression, brightness, and contrast are directly related to the visibility of manipulation artifacts. Using these features for routing can therefore provide a more task-specific form of adaptive inference.

In Summary, existing work has addressed low-resolution robustness, frequency-domain modeling, lightweight architectures, and adaptive inference, but these directions are usually studied separately. Low-resolution methods improve robustness but often do not reduce inference cost; lightweight methods improve efficiency but rarely adapt to input quality; and multi-branch frequency models improve representation strength but may increase complexity. AdaGate-DF addresses this gap by combining RGB and DCT representations with a quality-aware gated mechanism that routes samples through fast, medium, or full inference paths based on image-quality features.

**Algorithm 1** Overview of Quality-Aware Gated Deepfake Detection Training Pipeline

**Require:** Dataset $D = \{(x_i, y_i)\}_{i=1}^{N}$, resolutions $r \in R$

**Ensure:** Predicted probability $\hat{y} \in [0,1]$

1: Initialize model parameters $\Theta$ and gate parameters $\Phi$
2: **for** each training epoch **do**
3:     **for** each mini-batch $(x, y) \subset D$ **do**
4:         $r \sim R$
5:         $x \leftarrow \textsc{Resize}(x, r)$; $x \leftarrow \textsc{Normalize}(x)$
6:         $x_{rgb} \leftarrow x$
7:         $x_{freq} \leftarrow \text{DCT}(\textsc{Gray}(x))$
8:         $q \leftarrow \textsc{QualityFeatures}(x)$
9:         $g^* \leftarrow \textsc{RouteTarget}(q)$
10:         $\{z^1_{rgb}, z^2_{rgb}, z^3_{rgb}\} \leftarrow E_{rgb}(x_{rgb})$
11:         $\{z^1_{freq}, z^2_{freq}, z^3_{freq}\} \leftarrow E_{freq}(x_{freq})$
12:         $z_{fused} \leftarrow \textsc{Fuse}(z^3_{rgb}, z^3_{freq})$
13:         $\hat{g} \leftarrow G(q)$
14:         $L_{cls} \leftarrow \sum_{k=1}^{3} \ell(z^k_{rgb}, y) + \ell(z^k_{freq}, y) + \ell(z_{fused}, y)$
15:         $L_{gate} \leftarrow \ell(\hat{g}, g^*)$
16:         $L_{total} \leftarrow L_{cls} + L_{gate}$
17:         Update $\Theta, \Phi$ using $L_{total}$
18:     **end for**
19: **end for**
20: **return** trained model

## III. Methodology

AdaGate-DF uses a dual branch architecture that combines spatial features and frequency features derived using DCT for a deeper analysis. An adaptive gating module is added to utilize the image-quality cues to select a dynamic inference path, allowing the model to adjust computation based on input quality. The following subsections describe the preprocessing, feature extraction, gating, and training procedure. The entire architecture flow is shown in the Figure 1.

### A. Dataset and Preprocessing

We consider a supervised deepfake detection setting, where each input image $x$ is associated with a binary label $y \in \{0, 1\}$ indicating real or fake. To simulate real-world conditions, we incorporate multi-resolution inputs during training. Specifically, a target resolution $r$ is randomly sampled from a predefined set $R \in \{128 \times 128, 224 \times 224, 256 \times 256, 384 \times 384\}$, and each image is resized accordingly. This enables the model to learn robustness to varying input resolution.

All images are normalized before being passed into the network. In addition to the standard RGB representation, a frequency-domain representation is generated by applying DCT on the grayscale images. This produces complementary inputs that capture both spatial and frequency-based artifacts.

### B. Quality-Feature Estimation and Quality-Aware Adaptive Routing

To enable adaptive computation, a set of lightweight quality features is extracted from each input image. These features capture low-level image characteristics, including resolution, sharpness, blur, compression artifacts, brightness, and contrast. The resulting feature vector is denoted as $q$.

A lightweight gating network $G(\cdot)$ is used to map the quality feature vector $q$ to a routing decision $\hat{g} = G(q)$. This routing decision determines the level of computation required for processing the input. The gating mechanism allows the model to dynamically adjust its computational effort based on input quality, enabling efficient processing of high-quality samples while reserving deeper computation for more challenging inputs.

### C. Dual-Brach Gated Architecture

The proposed framework employs a dual-branch architecture consisting of a spatial and a frequency branch, with multiple exits. The spatial branch processes RGB information, while the frequency branch operates on DCT-transformed inputs to capture compression and manipulation artifacts. Each branch is implemented as a convolutional encoder with three exit points. These exits allow the network to produce predictions at different depths, corresponding to shallow, intermediate, and deep representations. Shallow layers capture low-level patterns, while deeper layers encode more complex and abstract forgery features. This multi-exit architecture enables the model to avoid unnecessary computation by selecting an appropriate inference path based on the input characteristics.

The proposed framework incorporates a multi-exit design that supports three inference paths corresponding to different computational budgets.

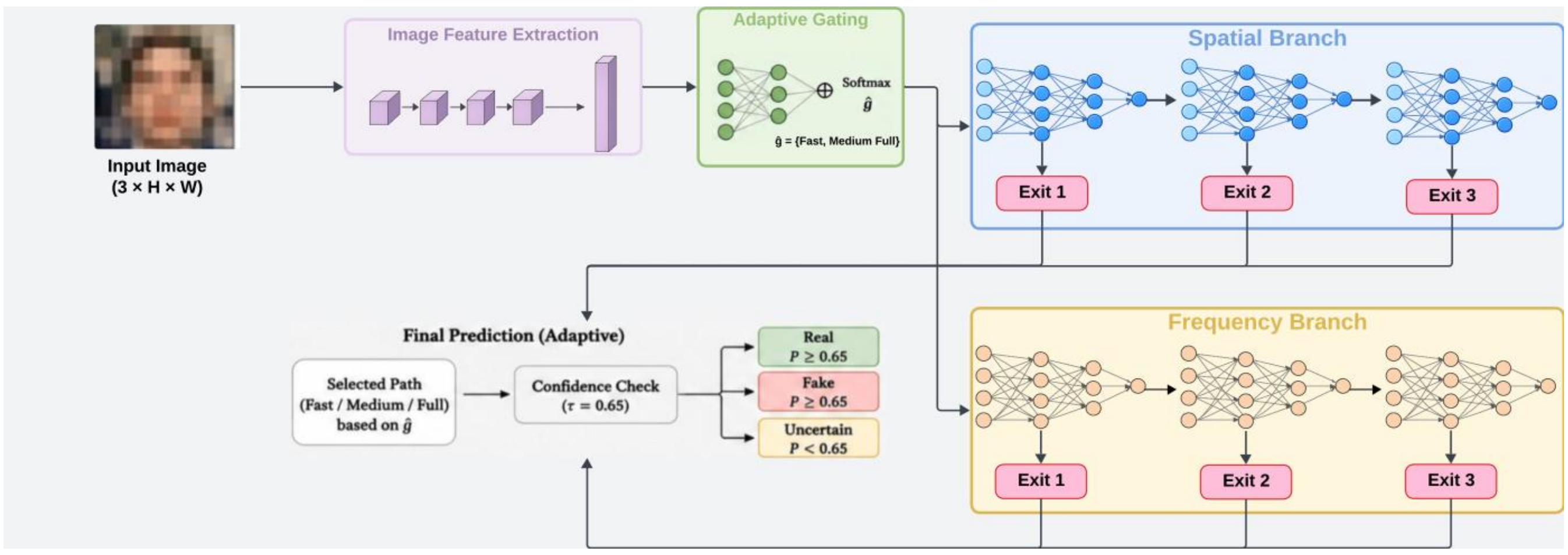

Fig. 1: Architecture Diagram of AdaGate-DF

- The **Fast path** utilizes shallow features from both the RGB and frequency branches to generate predictions with minimal computational cost. This path is primarily used for high-quality or less complex inputs.
- The **Medium path** leverages intermediate features to provide a balance between computational efficiency and detection performance.
- The **Full path** utilizes the deepest feature representations from both branches. These features are fused and passed through a final classifier to produce the most accurate prediction, particularly for low-quality or challenging inputs.

### D. Training and Inference Strategy

During training, all branches and exit points are optimized jointly. Each exit produces a prediction, and a classification loss is applied to all outputs to ensure that meaningful representations are learned at different depths. The entire training pipeline is explained in Algorithm 1.

The gating network is trained using supervision derived from quality-based routing rules, which assign target paths based on the input characteristics. This encourages the model to route simpler inputs to shallow paths and more complex inputs to deeper paths. The overall training objective consists of a combination of classification losses from all exits and a gating loss for the routing decision. This joint optimization allows both the detection model and the gating mechanism to be learned simultaneously.

During inference, the quality feature vector $q$ is first computed and passed through the gating network to obtain the routing decision $g\hat{}$. Based on this decision, the input is processed through the selected path (fast, medium, or full). This adaptive inference strategy reduces computational cost for simpler inputs while maintaining strong detection performance for more challenging cases.

## IV. Experimental Setup

We conduct experiments across multiple configurations, to evaluate performance from multiple perspectives: uncertainty-aware performance on the held out dataset (test), robustness to input resolution, and computational efficiency. We include both balanced and imbalanced dataset conditions, multiple input resolutions, and timing measurements for training and inference. Since the state-of-the-art models did not include uncertainty in their original implementation, we added it in our re-implementation for fair comparison between the models.

TABLE I: Summary of Experimental Configuration

| Setting | Value |
|---|---|
| Datasets | Celeb-DF, FF++ |
| Input resolutions | 128, 224, 256, 384 |
| Base inference resolution | 224×224 |
| Batch size | 32 |
| Random seed | 42 |
| AdaGate-DF optimizer | AdamW |
| AdaGate-DF learning rate | $2 \times 10^{-4}$ |
| AdaGate-DF weight decay | $1 \times 10^{-4}$ |
| AdaGate-DF epochs | 15 |
| FF++ frame extraction rate | 2 FPS |
| FF++ max frames per video | 8 |

### A. Dataset split and configuration

The experiments were conducted on two widely used deepfake detection benchmarks: Celeb-DF [2] and FF++ [1]. These datasets provide a controlled evaluation setting for analyzing detection performance under different input resolutions while minimizing the influence of background context. In our processed split, Celeb-DF contains 80,824 training images, 10,104 validation images, and 10,103 testing images, with approximately balanced real and fake classes.

FF++ is used to evaluate model behavior under a more naturally imbalanced setting. We extract image frames from videos containing real samples and multiple manipulation types. The dataset contains 7,000 videos, including 1,000 real videos and 6,000 manipulated videos, resulting in an approximate 1:6 real-to-fake ratio. Frames are extracted at 2 FPS, with a maximum of 8 frames per video, producing 55,953 total frames. The resulting frame-level split contains 44,756 training frames, 5,600 validation frames, and 5,597 testing frames. To prevent frame-level leakage, the dataset is split at the video level using an 80/10/10 train, validation, and test partition.

### B. Model Configuration

All images are resized before feature extraction and model evaluation. To evaluate performance under varying image-quality conditions, we test all models at four input resolutions: $128 \times 128$, $224 \times 224$, $256 \times 256$, and $384 \times 384$. These resolutions are selected to represent low-resolution, standard-resolution, and higher-resolution input settings. The $224 \times 224$ setting is used as the base resolution for the main performance and inference-time comparisons, while the remaining resolutions are used to evaluate robustness to resolution changes.

Images are normalized using ImageNet mean and standard deviation values. For AdaGate-DF, each input is represented using two complementary branches: an RGB spatial branch and a frequency branch. The frequency branch is generated by converting the input image to grayscale and applying DCT. In addition, lightweight image-quality features are extracted from each sample, including resolution, blur, sharpness, compression proxy, brightness, and contrast. These quality features are used by the gating module to select the appropriate inference path.

AdaGate-DF is trained using the AdamW optimizer with a learning rate of $2 \times 10^{-4}$, weight decay of $1 \times 10^{-4}$, batch size 32, and 15 training epochs. The complete experimental configurations is summarized in Table I. The model is trained using a combined loss that includes classification losses from the RGB branch, frequency branch, fused prediction output, and a gate-routing loss. The adaptive model is evaluated in four configurations: the full gated model, fast-only exit, medium-only exit, and full-only exit. The fixed-exit variants are included to measure the contribution of the adaptive gating mechanism. The combined loss function is shown in Equation 1, where $L_{rgb}$ and $L_{dct}$ supervise the RGB and frequency branches, $L_{fused}$ supervises the fused prediction, $L_{gate}$ supervises the routing decision, and $\lambda$ controls the contribution of the gate loss.

$$L_{total} = L_{rgb} + L_{dct} + L_{fused} + \lambda L_{gate} \tag{1}$$

All experiments were implemented in PyTorch and executed on the CSU TIDE high-performance computing infrastructure [29]. Training and evaluation were performed using GPU-accelerated compute nodes equipped with NVIDIA L40 GPUs, each with 48 GB of GPU memory. The same computing environment was used for the proposed method and the compared models to ensure that training-time and inference-latency comparisons were measured under consistent hardware conditions.

### C. Evaluation Metrics

We evaluate all models using accuracy, AUC, precision, recall, F1-score, uncertainty rate, training time, and inference latency. AUC is particularly important for the FF++ evaluation because the dataset split is highly imbalanced and accuracy-based metrics can be influenced by the majority class.

In addition to standard classification metrics, we report uncertainty rate to measure how often a model produces a low-confidence prediction. For AdaGate-DF, a prediction is marked uncertain when the maximum predicted class confidence is below the threshold $\tau = 0.65$. For probability-based models, predictions near the decision boundary are treated as uncertain using the interval $[0.4, 0.6]$. This protocol allows uncertainty to be estimated across models with different output formats while preserving a common interpretation of low-confidence predictions. The uncertainty rate is computed as the percentage of test samples marked uncertain:

$$UR = \frac{N_{uncertain}}{N_{total}} \times 100. \tag{2}$$

### D. State-of-the-Art Model Comparison

We compare AdaGate-DF with three state-of-the-art efficient deepfake detection models: MaD-CoRN [19], DefakeHop++ [28], and ShuffleNetV2 [27]. These models were selected because they represent different lightweight detection strategies relevant to low-resource deepfake detection.

MaD-CoRN is a lightweight deepfake detection approach based on convolutional reservoir networks, while DefakeHop++ is an enhanced lightweight detector that uses handcrafted feature extraction with classifier-based prediction. ShuffleNetV2 is included as an efficient CNN architecture designed for low-latency inference.

## V. Results and Discussion

This section discusses the results observed across all our experiments organized into three subsections. First we discuss the uncertainty-aware performance for all the models, followed by the resolution based performance for more generalized analysis. Later, we analyze the training and inference latency by comparing the time taken by each model to provide results.

### A. Uncertainty-Aware Performance on Celeb-DF and FF++ datasets

Table II compares the proposed gated model with the state-of-the-art methods on Celeb-DF and FF++. On Celeb-DF, the proposed gated model achieves an AUC of 0.9370, and an F1-score of 0.8654, performing better than two of the three compared models, MaD-CoRN and DefakeHop++. ShuffleNetV2 obtains the highest overall classification performance on Celeb-DF; however, the proposed method remains competitive while also providing adaptive routing and uncertainty-aware prediction. The benefit of the proposed architecture is further highlighted in the training and inference time comparison, explained in Section V-C.

TABLE II: Test dataset performance on Celeb-DF and FF++. U_Rate denotes the percentage of samples marked uncertain under the evaluation protocol.

| Model | Celeb-DF | | | | | | FF++ | | | | | |
|---|---|---|---|---|---|---|---|---|---|---|---|---|
| | Acc | AUC | Prec | Rec | F1 | U_Rate | Acc | AUC | Prec | Rec | F1 | U_Rate |
| Gated Dual Branch | 0.8637 | 0.9370 | 0.8572 | 0.8737 | 0.8654 | 22.23 | 0.8571 | 0.5163 | 0.8571 | 1.0000 | 0.9230 | 0.00 |
| Fast Only | 0.6535 | 0.7114 | 0.6478 | 0.6774 | 0.6622 | 83.31 | 0.8571 | 0.5295 | 0.8571 | 1.0000 | 0.9230 | 0.00 |
| Medium Only | 0.8205 | 0.9047 | 0.8164 | 0.8285 | 0.8224 | 37.25 | 0.8571 | 0.5101 | 0.8571 | 1.0000 | 0.9230 | 0.00 |
| Full Only | 0.9088 | 0.9709 | 0.8941 | 0.9280 | 0.9107 | 6.61 | 0.8571 | 0.4182 | 0.8571 | 1.0000 | 0.9230 | 0.00 |
| MaD-CoRN | 0.5780 | 0.6096 | 0.5782 | 0.5864 | 0.5823 | 85.01 | 0.8571 | 0.5592 | 0.8571 | 1.0000 | 0.9230 | 0.00 |
| DefakeHop++ | 0.8002 | 0.8831 | 0.7858 | 0.8269 | 0.8058 | 20.25 | 0.8531 | 0.2331 | 0.8565 | 0.9954 | 0.9207 | 1.55 |
| ShuffleNetV2 | 0.9876 | 0.9993 | 0.9858 | 0.9895 | 0.9877 | 0.54 | 0.8571 | 0.5693 | 0.8571 | 1.0000 | 0.9230 | 0.27 |

We also conducted further experiments to analyze if having the gated approach results in performance changes. The gated model substantially improves over the fast-only and medium-only configurations, indicating that relying on a single shallow or intermediate representation is insufficient for many samples. The full-only model achieves the strongest performance among the proposed variants, but it applies the deepest computation to every input. In contrast, the gated model provides a more deployment-oriented tradeoff by adapting the inference path according to input quality. Therefore while the gated approach leads to slight performance degradation compared to the forced binary decision, it offers a faster training pipeline and makes it generalized for all the images instead of one particular image type.

Uncertainty rate provides additional insight beyond standard classification metrics. On Celeb-DF, the fast-only exit has a high uncertainty rate of 83.31%, while the gated model reduces this value to 22.23%, showing that adaptive routing helps reduce low-confidence predictions by allowing more difficult samples to receive deeper analysis. However, uncertainty must be interpreted together with AUC and F1-score, since a low uncertainty rate alone does not necessarily indicate strong discriminative performance. Compared with MaD-CoRN, AdaGate-DF achieves substantially higher AUC and F1-score on Celeb-DF while reducing the uncertainty rate from 85.01% to 22.23%. This suggests that the proposed adaptive spatial-frequency design produces more reliable predictions under this evaluation setting.

To evaluate model behavior under a more realistic class distribution, we retained the imbalanced FF++ setting, which contains an approximate 1:6 ratio of real to fake samples. Because this imbalance can inflate threshold-dependent metrics such as accuracy, precision, recall, and F1-score, we use AUC as the primary comparison metric for this dataset. Although several models achieve similar accuracy and F1-score values, their low AUC scores indicate limited class separation and suggest bias toward the majority class. ShuffleNetV2 obtains the highest AUC of 0.5693, followed closely by MaD-CoRN, while the proposed gated model achieves an AUC of 0.5163 and outperforms DefakeHop++ on this metric. These results show that the proposed method is affected by the imbalanced FF++ setting, but it remains competitive with the evaluated models and performs more more stable than DefakeHop++ across the tested resolutions. The near-zero uncertainty rates observed for several models further indicate that confidence alone is insufficient under class imbalance, since models may produce confident predictions despite weak discriminative performance. Therefore, the FF++ results should be interpreted cautiously and supplemented with balanced accuracy, specificity, false positive rate, false negative rate, and confusion-matrix analysis, which will be addressed in the future work.

TABLE III: Resolution-Based Accuracy and AUC Across Models on Celeb-DF and FF++

| Model | Resolution | Celeb-DF | | FF++ | |
|---|---|---|---|---|---|
| | | Accuracy | AUC | Accuracy | AUC |
| Gated Dual Branch | 128 | 0.8112 | 0.8954 | 0.8571 | 0.5043 |
| | 224 | 0.8638 | 0.9364 | 0.8571 | 0.5253 |
| | 256 | 0.8689 | 0.9400 | 0.8571 | 0.5214 |
| | 384 | 0.9108 | 0.9708 | 0.8571 | 0.5202 |
| MaD-CoRN | 128 | 0.5582 | 0.6085 | 0.8571 | 0.5639 |
| | 224 | 0.5900 | 0.6250 | 0.8571 | 0.5569 |
| | 256 | 0.5898 | 0.6243 | 0.8571 | 0.5590 |
| | 384 | 0.5741 | 0.5987 | 0.8571 | 0.5575 |
| DefakeHop++ | 128 | 0.5100 | 0.7711 | 0.8565 | 0.4883 |
| | 224 | 0.8002 | 0.8831 | 0.8531 | 0.2331 |
| | 256 | 0.7695 | 0.8700 | 0.8569 | 0.3304 |
| | 384 | 0.6324 | 0.7896 | 0.8571 | 0.4094 |
| ShuffleNetV2 | 128 | 0.5716 | 0.8731 | 0.8571 | 0.5451 |
| | 224 | 0.9876 | 0.9993 | 0.8571 | 0.5836 |
| | 256 | 0.9840 | 0.9989 | 0.8571 | 0.5717 |
| | 384 | 0.8927 | 0.9742 | 0.8571 | 0.5765 |

## B. Resolution-Based Evaluation

Resolution-based testing was conducted to evaluate how each model responds to changes in input quality. This is important for real-world deepfake detection because images and video frames are often compressed, resized, downsampled, or captured at inconsistent resolutions. Since the proposed gated framework uses quality-related features to determine the inference path, testing across multiple resolutions allows us to examine whether the model remains reliable when the amount of available visual detail changes.

We evaluate all models at 128×128, 224×224, 256×256, and 384×384 resolutions. These settings represent low-resolution, standard-resolution, and higher-resolution input conditions. Lower resolutions test whether the models can still identify manipulation cues when fine-grained facial artifacts are reduced, while higher resolutions evaluate whether additional spatial and frequency information improves detection. However, higher resolution does not always guarantee better performance, since resizing and upscaling can introduce interpolation artifacts or alter high-frequency patterns. Therefore, $384 \times 384$ is used as the high-resolution evaluation setting.

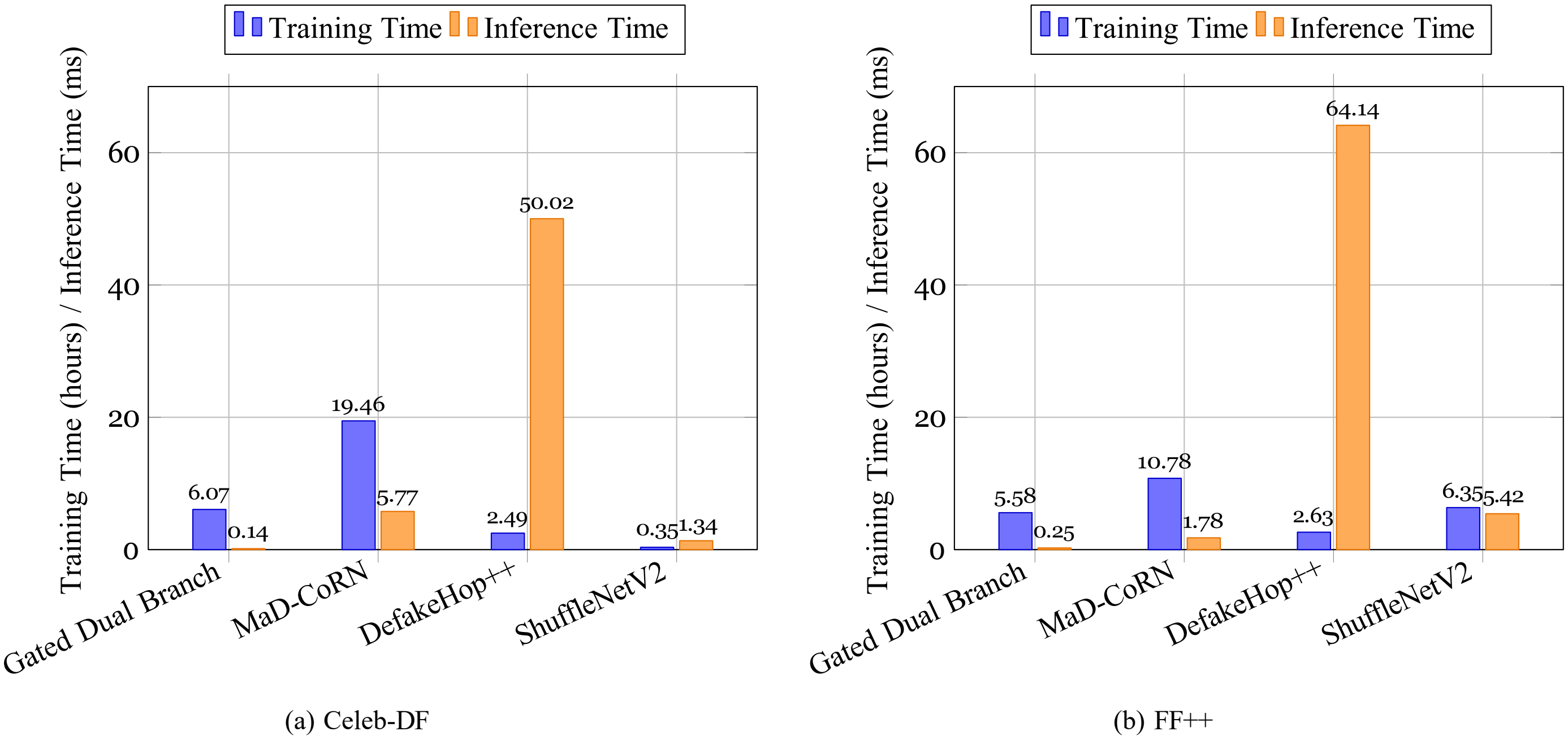

Fig. 2: Training (hrs) and inference (ms) time comparison across evaluated models. The shared y-axis is used only for compact visualization; the bars should be interpreted using their respective units.

As shown in Table III, the proposed gated model shows a consistent improvement on Celeb-DF as resolution increases, with AUC improving from 0.8954 at 128×128 to 0.9708 at 384×384, indicating the proposed spatial-frequency representation benefits from additional image detail while still maintaining strong performance at lower resolution. Compared with the SOTA models, the proposed method performs more consistently than MaD-CoRN and DefakeHop++ across the tested resolutions. ShuffleNetV2 achieves the highest AUC at 224×224 and 256×256, but its performance decreases at 384×384, showing that higher resolution does not uniformly improve all architectures.

For FF++, the high class imbalance leads to consistently high accuracy values across resolutions, while AUC remains low for most models. Even under this setting, the proposed gated model shows more stable behavior than DefakeHop++ across all tested resolutions. Specifically, the proposed method achieves AUC values between 0.5043 and 0.5253, whereas DefakeHop++ varies more substantially from 0.2331 to 0.4883. This indicates that although the proposed model is affected by the imbalanced evaluation setting, it maintains more consistent class-separation behavior across resolutions than DefakeHop++. ShuffleNetV2 and MaD-CoRN obtain slightly higher AUC values on FF++, but the gated model remains competitive while also providing adaptive routing and uncertainty-aware prediction.

### C. *Training and Inference Efficiency*

In addition to classification performance, we evaluate the computational efficiency of the proposed method and the compared models. Figure 2 reports the training and inference efficiency of the evaluated models on Celeb-DF and FF++. Training time is reported in hours, while inference time is reported in milliseconds per image at 224×224 resolution. The 224×224 setting is used as the base inference resolution because it provides a common evaluation point across all compared models and corresponds to a standard input size used in many image-based detection pipelines.

The results show that training time and inference time do not always follow the same trend. For example, DefakeHop++ has relatively low training time, but it has the highest inference latency on both datasets. This makes it less suitable for high-throughput deployment settings where many images or frames must be processed. In contrast, MaD-CoRN requires the highest training time on both datasets and also has higher inference latency than the proposed method. The proposed gated model provides a more favorable efficiency profile, requiring 6.07 hours of training and 0.14 ms inference time on Celeb-DF, and 5.58 hours of training and 0.25 ms inference time on FF++.

These results support the deployment motivation of the proposed approach. Although ShuffleNetV2 has the lowest training time on Celeb-DF, the proposed gated model achieves lower inference latency while also providing adaptive routing and uncertainty-aware prediction. On FF++, the proposed model has the lowest inference latency among the compared methods. This indicates that the gated architecture is effective not only as a detection model, but also as a practical inference framework for resource-constrained deepfake detection.

## VI. Conclusion

This paper presented **AdaGate-DF**, an adaptive gated deepfake detection framework designed for low-resolution and resource-constrained environments. The proposed method combines RGB spatial representations with DCT-based frequency representations and uses lightweight image-quality features to route samples through fast, medium, or full inference paths. The results show that AdaGate-DF provides a strong balance between detection performance, uncertainty-aware prediction, and computational efficiency. On Celeb-DF, the proposed method outperforms MaD-CoRN and DefakeHop++ while maintaining low inference latency, and the resolution-based results show consistent improvement as input resolution increases. These findings support the value of quality-aware adaptive routing for practical deepfake detection under variable image-quality conditions. The FF++ results further highlight the importance of AUC and class-balanced analysis when evaluating under imbalanced dataset conditions.

Future work will extend this evaluation by incorporating additional class-balanced metrics, including balanced accuracy, specificity, false positive rate, false negative rate, and confusion matrices, particularly for the imbalanced dataset. We will also evaluate timing under a unified hardware protocol and analyze the percentage of samples assigned to each gate across different resolutions. This would provide a more detailed understanding of how the adaptive routing mechanism behaves under changing input quality. Additional extensions may include improved gate supervision, and video-level temporal modeling.